%% file: ex_article.tex
\documentclass[hidelinks,onefignum,onetabnum]{siamart250211}

\input{ex_shared}

\ifpdf
\hypersetup{
  pdftitle={An Efficient Conditional Score-based Filter for High Dimensional Nonlinear Filtering Problems},
  pdfauthor={Zhijun Zeng, Weiye Gan, Junqing Chen, and Zuoqiang Shi}
}
\fi

\usepackage{amssymb}

\begin{document}

\maketitle

% REQUIRED
\begin{abstract}
In many engineering and applied science domains, high-dimensional nonlinear filtering is still a challenging problem. Recent advances in score-based diffusion models offer a promising alternative for posterior sampling but require repeated retraining to track evolving priors, which is impractical in high dimensions. In this work, we propose the Conditional Score-based Filter (CSF), a novel algorithm that leverages a set-transformer encoder and a conditional diffusion model to achieve efficient and accurate posterior sampling without retraining. By decoupling prior modeling and posterior sampling into offline and online stages, CSF enables scalable score-based filtering across diverse nonlinear systems. Extensive experiments on benchmark problems show that CSF achieves superior accuracy, robustness, and efficiency across diverse nonlinear filtering scenarios.
\end{abstract}

% REQUIRED

% REQUIRED

\section{Introduction}
Filtering is a fundamental problem in state estimation for stochastic dynamical systems, with widespread applications in robot vision\cite{chen2011kalman,thrun2002particle,chen2011kalman}, weather forecasting\cite{delle2011kalman,carrassi2018data}, signal processing\cite{arce2004nonlinear,karimi2002nonlinear}, and beyond. For linear Gaussian state-space models, the Kalman filter affords an analytic, minimum-variance solution for the latent state\cite{kalman1961new}. Nonlinear filtering, in which the state transition and/or observation model is nonlinear, has emerged as a significant research area, motivating a broad spectrum of methods to address challenges in autonomous systems\cite{brunke2001nonlinear,rigatos2012nonlinear,cui2018path}, materials science\cite{archibald2018backward}, biology\cite{cogan2021data}, finance\cite{javaheri2003filtering,frey2001nonlinear}, and other domains. The primary objective of filtering is to sequentially integrate noisy, partial observations in order to infer the underlying unobservable state of interest with maximal accuracy and computational efficiency. In this paper, we consider the following nonlinear filtering problem:
\begin{equation}\label{eqn:nfp}
\begin{array}{cccl}
     \text{State}:& \mathbf{x}_{k+1} &=&f(\mathbf{x}_{k}) + \mathbf{v}_k  \\
     \text{Observation}:&y_{k+1} &=&h(\mathbf{x}_{k+1})  + \mathbf{w}_k 
\end{array},
\end{equation}
where $\mathbf{x}_{k}\in \mathbb{R}^d$ is a vector of states at step $k$, $\mathbf{y}_k\in \mathbb{R}^m$ is a vector of observations at step $k$, $f(\mathbf{x})\in \mathbb{R}^d, h(\mathbf{x})\in \mathbb{R}^m$ are the known nonlinear drift term and observation function, respectively, and $\mathbf{v}_k,\mathbf{w}_k$ are random variables that follows given probability law representing the uncertainty of the state evolution and observation. For the filtering problem, one have to give an accurate and instantaneous estimate of $\mathbf{x}_{k}$ throught the noisy observation $\mathcal{Y}_{k} \triangleq \{y_s :0\leq s \leq k\}$. In practical applications, only discrete-time noisy measurements $\widehat{\mathcal{Y}}_k \triangleq \{\mathbf{y}_{i}\}_{i=1}^{k}$
are available for state estimation, and one should find the best estimate for the conditional expectation $\hat{\mathbf{x}}_k:\mathbb{E}[\mathbf{x}_k|\widehat{\mathcal{Y}}_k]$, which which requires an accurate characterization of the posterior distribution $q\bigl(\mathbf{x}_k \mid \widehat{\mathcal{Y}}_k\bigr).$ 

To approximate the posterior distribution \(q(\mathbf{x}_k\mid\widehat{\mathcal{Y}}_k)\) in nonlinear filtering, a common approach is to invoke a Gaussian approximation and propagate its conditional mean and covariance.  Numerous algorithms of this class include the extended Kalman filter (EKF)\cite{jazwinski2013stochastic}, the unscented Kalman filter (UKF)\cite{julier2002new,julier2004unscented}, and the ensemble Kalman filter (EnKF)\cite{evensen2003ensemble}.  As linearization‐based schemes, these methods perform satisfactorily when the system nonlinearity is moderate  or the noise levels are sufficiently low.  However, in  high-level or non-Gaussian noise regimes, Kalman‐type filters can diverge and generally lack rigorous convergence guarantees. Another widely used approach is the particle filter method\cite{djuric2003particle, andrieu2010particle}. Also known as sequential Monte Carlo, this method represents the posterior density \(q(\mathbf{x}_k\mid\widehat{\mathcal{Y}}_k)\) by an ensemble of weighted particles. At each time step, particles are propagated through the nonlinear state dynamics, assigned importance weights via the observation likelihood, and then resampled to concentrate computational effort on high-probability regions. In the asymptotic regime \(N\to\infty\), the empirical distribution of particles converges almost surely to the true posterior irrespective of the noise statistics\cite{crisan2002survey}. However, particle filters suffer from the curse of dimensionality: as the state and observation dimensions increase, the variance of importance weights grows and the number of particles required for a given accuracy scales exponentially with problem dimension, rendering high-dimensional applications computationally prohibitive.

Over the past decade, score-based diffusion models have emerged as a powerful framework for learning high-dimensional data distributions by approximating the Stein score, \(\nabla_{\mathbf{x}}\log p(\mathbf{x})\)\cite{song2020score,rombach2022high,croitoru2023diffusion}.  It has been widely recognized that diffusion-based priors can be leveraged for posterior sampling to solve inverse problems—both linear and nonlinear—including image reconstruction \cite{chung2023diffusion,graikos2022diffusion,zhu2023denoising,song2023loss,chung2022improving,songsolving}and differential-equation-constrained inverse problem\cite{jiang2025ode,li2024learning}.  However, in nonlinear filtering the state dynamics induce a time-varying prior \(q(\mathbf{x}_k)\), which must be modeled separately at each step \(k\).  This requirement makes existing diffusion-based filtering approaches\cite{bao2024score,ding2024nonlinear} computationally prohibitive in high-dimensional settings.  

In this paper, we present an efficient conditional score-based filter (CSF) that leverages a conditional diffusion model to enable training-free and accurate posterior distribution modelling in high-dimensional nonlinear filtering problems. In the offline stage, a set encoder extracts informative features from an ensemble of particles; these features condition a diffusion model pre-trained to approximate a variety of prior distributions. This design is justified by the ready availability of prior distributions in nonlinear filtering and the demonstrated efficacy of conditional diffusion model. In the online stage, at each time step CSF tracks the evolving prior distribution using a finite Monte Carlo ensemble, and employs these particles as conditions for diffusion-based posterior sampling to approximate $q\bigl(\mathbf{x}_k \mid \widehat{\mathcal{Y}}_k\bigr).$
We validate the proposed CSF on several high-dimensional benchmark problems, demonstrating superior estimation accuracy and computational efficiency. Our main contributions are twofold:
\begin{itemize}
    \item As a novel nonlinear filtering algorithm, our method accurately approximates high-dimensional posterior distributions with a  and exhibits robust performance in the presence of unknown external perturbations.
    \item By learning set-based particle features, we decouple prior distribution modeling and posterior sampling into distinct offline and online stages, thereby enabling fast score-based sampling to be applied to high-dimensional nonlinear filtering for the first time.  Moreover, the inherent generalization capability of the conditional diffusion model permits direct reuse for prior modeling across diverse nonlinear filtering scenarios without additional retraining.
\end{itemize}

The remainder of this paper is organized as follows. Section 2 presents the proposed conditional score-based filter. Section 3 reports a series of numerical experiments on nonlinear filtering problems ranging from low to high-dimensional regimes. Finally, Section 4 concludes the paper with a discussion of results and directions for future work.

\section{Method}
In this section, we present the key components of the proposed Conditional Score-based Filter (CSF). The methodology consists of two principal stages: an offline stage (Section~\ref{sec:offline}) and an online stage (Section~\ref{sec:online}).
\begin{figure}[htbp]\label{fig:architecture}
    \centering
        \includegraphics[width=\linewidth]{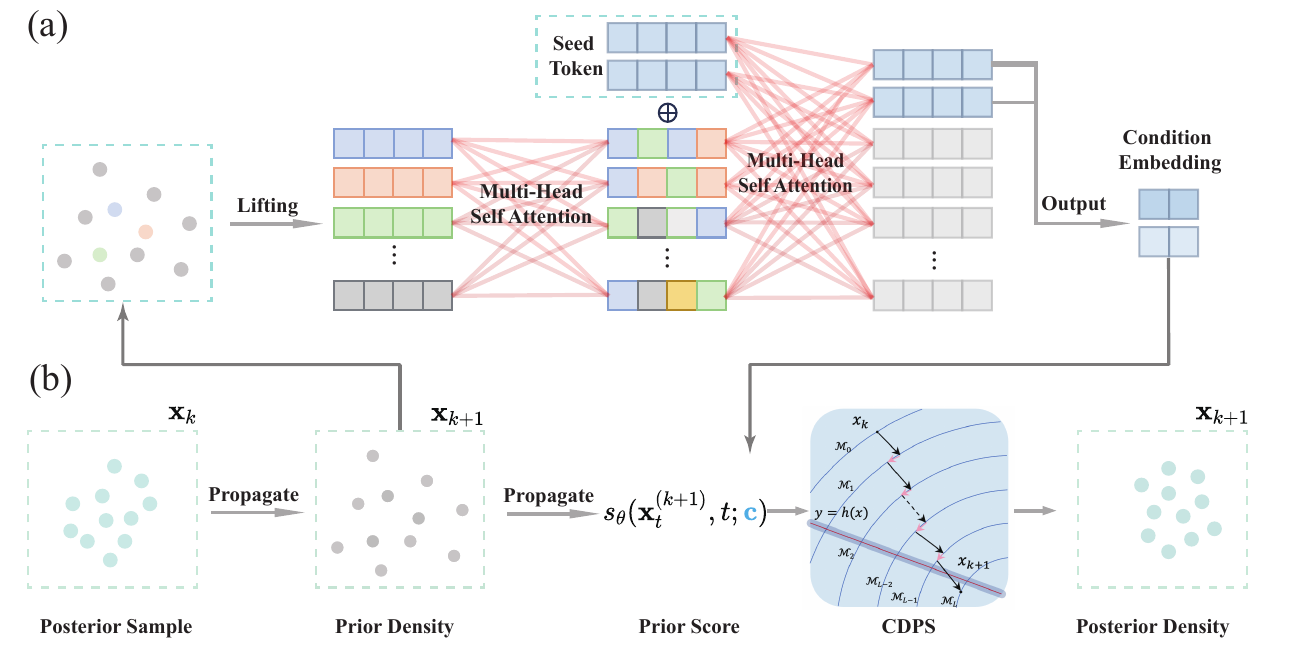}
    \caption{Schematic diagram of the Conditional Score-based Filter method. (a) The flowchart of the set transformer-based prior encoder; (b) The flowchart of the conditional diffusion posterior sampling procedure.}
    \label{figure:architecture}
\end{figure}
 \subsection{Offline Stage: Training Conditional Diffusion Models for Efficient Prior Distribution Modelling}\label{sec:offline}
\subsubsection{Set Transformer–based Prior Encoder}
In the sequential posterior-estimation framework of nonlinear filtering, the prior distribution \(q\) evolves at every discrete step \(k\).  Our objective is to train a single conditional generative model capable of representing the prior at any such step; conditioned on a compact embedding of this prior \(q\), the model can then draw posterior samples.  The only information available about the prior at step \(k\) is an ensemble of samples \(\{\mathbf{x}_i\}_{i=1}^{N}\subset\mathbb{R}^{d}\).  We therefore compress this ensemble into a neural statistic \(\mathbf{c}\) that must be (i) permutation-invariant and (ii) independent of the ensemble size \(N\) so that the generator can further infer the required posterior statistics from \(\mathbf{c}\) and the observation.  To satisfy both requirements we adopt the Set Transformer architecture\cite{lee2019set}, whose structure is illustrated in Fig.~\ref{fig:architecture}(a).

Let $\mathbf{u}_i$ be an input sample, a linear embedding then maps \(\mathbf{u}_i\) to a hidden state
\begin{equation}\label{eq:embed}
  \mathbf{h}_i^{(0)} = W_e\,\mathbf{u}_i + b_e,\quad i=1,\dots,N.
\end{equation}
To model interactions among samples, we stack \(L\)  Self‐Attention Blocks (SAB).  Denoting \(\mathbf{H}^{(0)}=[\mathbf{h}_1^{(0)};\dots;\mathbf{h}_N^{(0)}]\in\mathbb{R}^{N\times h}\), each block writes
\begin{equation}\label{eq:sab}
  \mathbf{H}^{(\ell)} 
  = \mathrm{LayerNorm}\Bigl(\mathbf{H}^{(\ell-1)} + \mathrm{MHA}\bigl(\mathbf{H}^{(\ell-1)},\mathbf{H}^{(\ell-1)},\mathbf{H}^{(\ell-1)}\bigr)\Bigr),
  \quad \ell=1,\dots,L.
\end{equation}
The multi-head attention (MHA) component is permutation-equivariant, a property formalized by
\begin{equation}\label{eqn:per_inv}
    \mathrm{MHA}(P\mathbf{Q},P\mathbf{K},P\mathbf{V}) = P\,\mathrm{MHA}(\mathbf{Q},\mathbf{K},\mathbf{V}),
\end{equation}
for any permutation matrix \(P\).This property is inherent to the attention mechanism, which operates on an unordered set of input vectors, ensuring the architecture is independent of the number of samples N and can therefore be applied to variable-sized datasets.

Finally, we aggregate the set into a single, permutation‐invariant embedding via Pooling by Multi‐Head Attention (PMA).  Introducing a learnable seed vector \(\mathbf{s}\in\mathbb{R}^h\), let
\(\widetilde{\mathbf{H}}=[\mathbf{s};\,\mathbf{H}^{(L)}]\in\mathbb{R}^{(N+1)\times h}\).  We then compute
\begin{equation}\label{eq:pma}
  \widetilde{\mathbf{Z}}
  = \mathrm{MHA}\bigl(\widetilde{\mathbf{H}},\widetilde{\mathbf{H}},\widetilde{\mathbf{H}}\bigr),
  \quad
  \mathbf{z} = \mathrm{LayerNorm}\bigl(\widetilde{\mathbf{Z}}_{1,:}\bigr)\in\mathbb{R}^h,
\end{equation}
where \(\mathbf{z}\) is invariant to any permutation of \(\{\mathbf{x}_i\}\) or changes in \(N\).  A final two‐layer MLP produces the conditional code:
\begin{equation}\label{eq:output}
  \mathbf{c} = W_2\,\rho\bigl(W_1\,\mathbf{z} + b_1\bigr) + b_2,
\end{equation}
where $\rho$ is the activation function. By construction, \(\psi_{\phi}:(\{\mathbf{x}_i\}_{i=1}^N)\mapsto\mathbf{c}\) is both permutation invariant and sample‐size independent, yielding an expressive neural statistic for downstream conditional diffusion.  

To help capture the statistics of a high-dimensional prior, we pre-compute the mean and variance of the samples
\begin{equation}\label{eq:stats}
  \boldsymbol\mu = \frac{1}{N}\sum_{i=1}^N \mathbf{x}_i,
  \quad
  \boldsymbol\sigma^2 = \frac{1}{N}\sum_{i=1}^N (\mathbf{x}_i - \boldsymbol\mu)^2,
\end{equation}
and augment each sample as
\begin{equation}\label{eq:augment}
  \mathbf{u}_i = \bigl[\mathbf{x}_i;\,\boldsymbol\mu;\,\boldsymbol\sigma^2\bigr]\in\mathbb{R}^{3d}.
\end{equation}
\subsubsection{Conditional Score-based Diffusion Models}
To enable generation of particles from a diverse set of prior distributions, we introduce conditional score-based diffusion models.  Score-based diffusion models provide a powerful framework for generating samples from complex, high-dimensional distributions by estimating the Stein score \(\nabla_{\mathbf{x}}\log p(\mathbf{x})\).  These models consist of two stochastic processes: a forward (noising) SDE that gradually perturbs clean data, and a reverse (denoising) SDE that reconstructs the data distribution. The forward process is defined by the Itô SDE:
\begin{equation}\label{eqn:forward_sde}
\left\{\begin{aligned}
    d\tilde{\mathbf{x}}_t &= f(t)\,dt + g(t)\,d\mathbf{w}_t,\\
    \tilde{\mathbf{x}}_0 &\sim p_{\mathrm{data}}(\mathbf{x}),
\end{aligned}\right.
\end{equation}
where \(\mathbf{w}_t\in\mathbb{R}^d\) is a standard Wiener process.  One can show that
\[
    p(\tilde{\mathbf{x}}_t\mid\tilde{\mathbf{x}}_0)
    = \mathcal{N}\bigl(\tilde{\mathbf{x}}_t \mid \alpha_t\,\tilde{\mathbf{x}}_0,\;\sigma_t^2\,\mathbf{I}\bigr),
\]
with
\begin{equation}\label{eqn:ab}
     \alpha_t = \exp\!\Bigl(\!\int_{0}^{t}f(s)\,ds\Bigr), 
    \quad
    \sigma_t^2 = \int_{0}^{t}\exp\!\Bigl(2\!\int_{\ell}^{t}f(s)\,ds\Bigr)\,g^2(\ell)\,d\ell,
\end{equation}
chosen such that \(\tilde{\mathbf{x}}_T\sim\mathcal{N}(0,\mathbf{I}_d)\).

The reverse (generative) process is described by the reverse-time SDE:
\begin{equation}\label{eqn:reverse_sde}
    d\tilde{\mathbf{x}}_t 
    = \bigl[f(t)\,\tilde{\mathbf{x}}_t \;-\; g^2(t)\,\nabla_{\tilde{\mathbf{x}}_t}\log p(\tilde{\mathbf{x}}_t)\bigr]\,dt 
    \;+\; g(t)\,d\bar{\mathbf{w}}_t,
\end{equation}
where \(\nabla_{\tilde{\mathbf{x}}_t}\log p(\tilde{\mathbf{x}}_t)\) is the time-dependent score, and \(d\bar{\mathbf{w}}_t\) denotes a backward-in-time Wiener increment.  In practice, the score is approximated by a neural network \(\mathbf{s}_\theta(\tilde{\mathbf{x}}_t,t)\) trained via denoising score matching:
\begin{equation}\label{eqn:score_matching}
    \theta^* 
    = \arg\min_\theta
    \mathbb{E}_{\substack{t\sim \mathcal{U}(\epsilon,T),\\
                            \tilde{\mathbf{x}}_0\sim p_{\mathrm{data}},\\
                            \tilde{\mathbf{x}}_t\sim p(\tilde{\mathbf{x}}_t\mid\tilde{\mathbf{x}}_0)}}
    \bigl\|\mathbf{s}_\theta(\tilde{\mathbf{x}}_t,t) - \nabla_{\tilde{\mathbf{x}}_t}\log p(\tilde{\mathbf{x}}_t\mid\tilde{\mathbf{x}}_0)\bigr\|_2^2.
\end{equation}

In nonlinear filtering, the prior \(p(\tilde{\mathbf{x}}_0)\) evolves over time according to the system dynamics.  Let \(\{\mathbf{x}_i\}_{i=1}^N\sim q(\mathbf{x})\) be an ensemble of particles drawn from a prior distribution at step $k$, and define a set encoder
\[
    \mathbf{c} = \psi_{\phi}\bigl(\{\mathbf{x}_i\}_{i=1}^N\bigr)
\]
that captures the underlying distributional features.  We then approximate the conditional score \(\nabla_{\tilde{\mathbf{x}}_t}\log p(\tilde{\mathbf{x}}_t\mid\mathbf{c})\) by a multi-input network \(\mathbf{s}_\theta(\tilde{\mathbf{x}}_t,t;\mathbf{c})\).  To train the conditional diffusion model, we jointly optimize the set encoder and score network by minimizing a conditional denoising score matching objective. Let \(p_{\mathrm{dist}}\) denote the meta-distribution over all priors of interest. The training objective is

\begin{equation}\label{eqn:cond_score_matching}
\begin{aligned}
    (\theta^*,\phi^*) &= \arg\min_{\theta,\phi}\; 
    &\mathbb{E}_{\substack{ t\sim \mathcal{U}(\epsilon,T),\\
                          \{\tilde{\mathbf{x}}_0,\{\mathbf{y}_i\}\}\sim p_{\mathrm{prior}},\\
                          p_{\mathrm{prior}}\sim p_{\mathrm{dist}},\\
                          \tilde{\mathbf{x}}_t\sim p(\tilde{\mathbf{x}}_t\mid\tilde{\mathbf{x}}_0,\mathbf{c})}}
    \bigl\|\mathbf{s}_\theta(\tilde{\mathbf{x}}_t,t;\mathbf{c}) 
               - \nabla_{\tilde{\mathbf{x}}_t}\log p(\tilde{\mathbf{x}}_t\mid\tilde{\mathbf{x}}_0;\mathbf{c})\bigr\|_2^2,\\
  \text{where }  \mathbf{c}& = \psi_{\phi}\bigl(\{\mathbf{x}_i\}_{i=1}^N\bigr).
\end{aligned}
\end{equation}
In practice, we approximate \(p_{\mathrm{dist}}\) by applying a standard particle filter to the target nonlinear filtering problem and recording the ensemble \(\{\mathbf{x}_i\}_{i=1}^N\) at each step \(k\) as an empirical prior.  Although particle filters are known to degrade in high-dimensional settings, a series of Monte Carlo simulations yield sufficiently diverse empirical priors to train the conditional diffusion model effectively. 

We refer to this algorithm as the offline stage of the our CSF method, and present a summary of the procedure in Algorithm \ref{alg:offline}
\begin{algorithm}[!t]
\caption{Offline Stage: Training Conditional Diffusion Model}
\label{alg:offline}
\begin{algorithmic}[1]
\REQUIRE Prior‐ensemble dataset 
  $\displaystyle \mathcal{D}=\Bigl\{\{\mathbf{x}_i^j\}_{i=1}^N\Bigr\}_{j=1}^M$, 
  learning rate $\eta$, epochs $E$, batch size $B$
\ENSURE Encoder params $\phi$, score‐net params $\theta$

\STATE Initialize $\phi,\theta$
\FOR{epoch = 1 to $E$}
  \FOR{each mini‐batch 
        $\{S^j\}_{j=1}^B\sim\mathcal{D}$, where $S^j=\{\mathbf{x}_i^j\}_{i=1}^N$}
    \STATE Uniformly sample $\displaystyle \mathbf{x}_0^j\sim S^j$, sample $t_j\sim\mathcal{U}(\epsilon,T)$ and $z_j\sim\mathcal{N}(0,I)$; \\[-.5ex]
           compute 
           $\alpha_{t_j},\sigma_{t_j}$ via closed‐form of Eq.~\eqref{eqn:ab}; 
    \STATE Set
           $\displaystyle\tilde{\mathbf{x}}_{t_j}^j
             = \alpha_{t_j}\,\mathbf{x}_0^j + \sigma_{t_j}\,z_j$
    \STATE Compute neural statistic 
      $\displaystyle \mathbf{c}^j = \psi_{\phi}(S^j)$
    \STATE Compute conditional score 
      $\displaystyle \hat{s}^j 
        = s_{\theta}(\tilde{\mathbf{x}}_{t_j}^j,\;t_j;\;\mathbf{c}^j)$
  \ENDFOR
  \STATE Compute loss:
    \[
      \mathcal{L}
        = \frac{1}{B}\sum_{j=1}^B
          \bigl\|\sigma_{t_j}\,\hat{s}^j + z_j\bigr\|_2^2
    \]
  \STATE Update $(\phi,\theta)\gets(\phi,\theta)-\eta\,\nabla_{\phi,\theta}\,\mathcal{L}$
\ENDFOR
\end{algorithmic}
\end{algorithm}
\subsection{Online Stage: Conditional Diffusion Posterior Sampling for Nonlinear Filtering}\label{sec:online}
Recall the nonlinear filtering problem in \eqref{eqn:nfp}, where the state \(\mathbf{x}_k\) evolves under the nonlinear map \(f\) with process noise \(\mathbf{v}_k\), and the measurements \(\mathbf{y}_k\) are produced by the observation map \(h\) with measurement noise \(\mathbf{w}_k\).
  The Bayesian filter framework is a paradigm widely adopted by methods such as the particle filter and EnKF to approximate the time-evolving posterior density \(q(\mathbf{x}_k \mid \widehat{\mathcal{Y}}_k)\).  At step \(k+1\), this framework proceeds in two steps: propagation of the prior and incorporation of the new measurement.

In the propagation step, the prior density at \(k+1\) is given by the Chapman–Kolmogorov integral
\begin{equation}
\label{eqn:ck}
\begin{array}{cc}
 \mathrm{Propagation:}    &  q(\mathbf{x}_{k+1}|\hat{\mathcal{Y}}_k)
=\int p\bigl(\mathbf{x}_{k+1}\mid \mathbf{x}_k\bigr)\,q_t(\mathbf{x}_k|\hat{\mathcal{Y}}_k)\,d\mathbf{x}_k,
\end{array}
\end{equation}
where $q(\mathbf{x}_{k+1}|\hat{\mathcal{Y}}_k)$ is called the prior distribution and  \(p(\mathbf{x}_{k+1}\mid \mathbf{x}_k)\) is the transition density induced by
$
 \mathbf{x}_{k+1} =f(\mathbf{x}_{k}) + \mathbf{v}_k.
$
Once the new observation \(\mathbf{y}_{t+\Delta t}\) arrives,
the posterior density $q(\mathbf{x}_{k+1}|\hat{\mathcal{Y}}_{k+1})$ is approximated by Bayes’ rule in the conditioning step:
\begin{equation}
\label{eqn:bayes_update}
\begin{array}{cc}
   \mathrm{Conditioning}:  & q(\mathbf{x}_{k+1}|\hat{\mathcal{Y}}_{k+1})
\;\propto\;
q(\mathbf{x}_{k+1}|\hat{\mathcal{Y}}_k)
\;q\bigl(\mathbf{y}_{k+1}\mid \mathbf{x}_{k+1}\bigr).
\end{array}
\end{equation}
The likelihood kernel $q\bigl(\mathbf{y}_{k+1}\mid \mathbf{x}_{k+1}\bigr)$ is determined by the observation operator $h(\cdot)$ and the noise distribution $\mathbf{w}_k$. Specifically,
\begin{equation}
q\bigl(\mathbf{y}_{k+1}\mid \mathbf{x}_{k+1}\bigr) ;\propto;\left\{
\begin{array}{ll}
\exp\left(-\tfrac{1}{2}\|\mathbf{y}_{k+1}-h(\mathbf{x}_{k+1})\|_{\Sigma^{-1}}^{2}\right) & \text{Gaussian noise}\\
 \left(1+\|\mathbf{y}_{k+1}-h(\mathbf{x}_{k+1})\|_{\Sigma^{-1}}^{2}\right)^{-\frac{d+1}{2}}&\text{Cauchy-type noise}.
\end{array}
\right.
\end{equation}
  Equations \eqref{eqn:ck} and \eqref{eqn:bayes_update} constitute the core of the Bayesian filter.  In high-dimensional, strongly nonlinear settings, accurately tracking the evolution of both prior and posterior distributions—and sampling from the resulting unnormalized kernel—poses significant computational challenges.  To address these, our CSF method employs a finite ensemble of particles propagated according to \eqref{eqn:nfp} to approximate the prior distribution, and leverages a pretrained conditional diffusion model to generate posterior samples in the conditioning step.  
\subsubsection{Propagation Step}
We adopt the particle filter paradigm to approximate both the prior and posterior densities via an ensemble of \(N\) particles.  At step \(k\), let \(\{\mathbf{x}_{k,i}\}_{i=1}^N\) denote the particle set that approximates the posterior density \(q(\mathbf{x}_k \mid \hat{\mathcal{Y}}_k)\) by the empirical measure
\begin{equation}\label{eqn:empirical_dist}
q(\mathbf{x}_k \mid \hat{\mathcal{Y}}_k)
\;\approx\;
\frac{1}{N}\sum_{i=1}^N \delta_{\mathbf{x}_{k,i}}(\mathbf{x}_k),
\end{equation}
where \(\delta\) is the Dirac delta.  In analogy with the standard particle filter, the prior density step \(k+1\) (cf.\ \eqref{eqn:ck}) is obtained by propagating each particle according to the state equation in \eqref{eqn:nfp}:
\begin{equation}\label{eqn:euler_maruyama}
\mathbf{x}_{k+1,i}
= \mathbf{x}_{k,i}
+ f\bigl(\mathbf{x}_{k,i}\bigr)
+\mathbf{v}_{k,i}.
\end{equation}

The pretrained set encoder \(\psi_{\phi}\) then extracts neural statistics $\mathbf{c}$ from the propagated ensemble:
\begin{equation}\label{eqn:set_encoder}
\mathbf{c}_{k+1}
= \psi_{\phi}\bigl(\{\mathbf{x}_{k+1,i}\}_{i=1}^N\bigr).
\end{equation}
These statistics condition the score network \(\mathbf{s}_{\theta}(\mathbf{x},t;\mathbf{c}_{k+1})\), which approximates the score function of the prior distribution in a denoising diffusion process.  Consequently, an arbitrary number of samples from the prior can be generated by integrating the reverse-time SDE:
\begin{equation}\label{eqn:prior_gen}
\left\{
\begin{array}{cl}
     d\tilde{\mathbf{x}}_t &= \bigl[f(t)\,\tilde{\mathbf{x}}_t \;-\; g^2(t)\mathbf{s}_{\theta}(\tilde{\mathbf{x}}_t,t;\mathbf{c}_{k+1})\,\bigr]\,dt 
    \;+\; g(t)\,d\bar{\mathbf{w}}_t  \\
   \tilde{\mathbf{x}}_T &  \sim\mathcal{N}(0,\mathbf{I}_d).
\end{array}\right.
\end{equation}
By approximating the full score function of the prior distribution rather than relying solely on the propagated particles, our CSF method has the potential to overcome the sample degeneracy and enables efficient posterior sampling even when the observation and prior supports are distant, particularly in high-dimensional settings where conventional particle filters fail.
\subsubsection{Conditioning via Conditional Diffusion Posterior Sampling}
In the propagation step, the learned score network 
\(\mathbf{s}_{\theta}(\tilde{\mathbf{x}},t;\mathbf{c}_{k+1})\) 
serves as a surrogate for the Stein score \(\nabla_{\tilde{\mathbf{x}}}\log p(\tilde{\mathbf{x}},t)\), 
whose marginal at \(t=0\) coincides with the prior \(q(\mathbf{x}_{t+\Delta t}\mid \hat{\mathcal{Y}}_t)\).  
In particle-based methods, the conditioning step aims to generate samples from the posterior 
\(q(\mathbf{x}_{k+1}\mid \hat{\mathcal{Y}}_{k+1})\).  
Analogous to \eqref{eqn:reverse_sde}, we obtain posterior samples by integrating the following conditional reverse-time SDE:
\begin{equation}\label{eqn:reverse_sde_cond}
\begin{aligned}
d\tilde{\mathbf{x}}_t 
&= \Bigl[f\bigl(\tilde{\mathbf{x}}_t,t\bigr)
- g^2(t)\,\nabla_{\tilde{\mathbf{x}}_t}\log p\bigl(\tilde{\mathbf{x}}_t\mid \hat{\mathcal{Y}}_{k+1}\bigr)\Bigr]dt
+ g(t)\,d\bar{\mathbf{w}}_t\\
&\approx \Bigl[f\bigl(\tilde{\mathbf{x}}_t,t\bigr)
- g^2(t)\,\nabla_{\tilde{\mathbf{x}}_t}\log p\bigl(\tilde{\mathbf{x}}_t\mid \mathbf{y}_{k+1}\bigr)\Bigr]dt
+ g(t)\,d\bar{\mathbf{w}}_t\,.
\end{aligned}
\end{equation}
By Bayes’ rule, the conditional score decomposes into two terms:
\begin{equation}\label{eqn_split_score}
\nabla_{\tilde{\mathbf{x}}_t}\log p\bigl(\tilde{\mathbf{x}}_t\mid \mathbf{y}_{k+1}\bigr)
= \nabla_{\tilde{\mathbf{x}}_t}\log p\bigl(\tilde{\mathbf{x}}_t\bigr)
+ \nabla_{\tilde{\mathbf{x}}_t}\log p\bigl(\mathbf{y}_{k+1}\mid \tilde{\mathbf{x}}_t\bigr).
\end{equation}
The first term is directly approximated by 
\(\mathbf{s}_{\theta}(\tilde{\mathbf{x}}_t,t;\mathbf{c}_{t+\Delta t})\), 
while the second term is intractable in general.  To retain computational efficiency, we approximate the likelihood score via Tweedie’s formula.  Under the Gaussian perturbation kernel
\[
p\bigl(\tilde{\mathbf{x}}_t\mid \tilde{\mathbf{x}}_0\bigr)
=\mathcal{N}\bigl(\tilde{\mathbf{x}}_t;\;\alpha_t\,\tilde{\mathbf{x}}_0,\;\sigma_t^2\mathbf{I}\bigr),
\]
Tweedie’s formula gives
\begin{equation}\label{eqn:tweedies}
\mathbb{E}\bigl[\tilde{\mathbf{x}}_0\mid \tilde{\mathbf{x}}_t\bigr]
= \frac{\tilde{\mathbf{x}}_t + \sigma_t^2\,\nabla_{\tilde{\mathbf{x}}_t}\log p(\tilde{\mathbf{x}}_t)}{\alpha_t}\,.
\end{equation}
Hence, following the idea of \cite{chung2023diffusion}, the likelihood score is approximated by
\begin{equation}\label{eqn:second_term}
\nabla_{\tilde{\mathbf{x}}_t}\log p\bigl(\mathbf{y}_{k+1}\mid \tilde{\mathbf{x}}_t\bigr)
\approx \nabla_{\tilde{\mathbf{x}}_t}\log p\bigl(\mathbf{y}_{k+1}\mid \hat{\mathbf{x}}_0(\tilde{\mathbf{x}}_t)\bigr),
\quad
\hat{\mathbf{x}}_0(\tilde{\mathbf{x}}_t)
= \mathbb{E}\bigl[\tilde{\mathbf{x}}_0\mid \tilde{\mathbf{x}}_t\bigr].
\end{equation}
Substituting these approximations into \eqref{eqn:reverse_sde_cond}, the complete conditional diffusion posterior‐sampling dynamics read
\begin{equation}\label{eqn:reverse_conditional_sde}
d\tilde{\mathbf{x}}_t
= \Bigl[f\bigl(\tilde{\mathbf{x}}_t,t\bigr)
- g^2(t)\Bigl(\mathbf{s}_{\theta}\bigl(\tilde{\mathbf{x}}_t,t;\mathbf{c}_{k+1}\bigr)
+ \nabla_{\tilde{\mathbf{x}}_t}\log p\bigl(\mathbf{y}_{k+1}\mid \hat{\mathbf{x}}_0(\tilde{\mathbf{x}}_t)\bigr)\Bigr)\Bigr]dt
+ g(t)\,d\bar{\mathbf{w}}_t.
\end{equation}
This conditional SDE enables efficient generation of posterior samples by combining the learned prior score with an approximate likelihood gradient, thereby mitigating sample degeneracy and achieving robust performance in high-dimensional filtering scenarios. The conditioning step is schematically illustrated in Fig~\ref{fig:architecture}(b). Furthermore, a concise summary of the online procedure is provided in Algorithm\ref{alg:online}.
\begin{algorithm}[!t]
\caption{Online Stage: Conditional Diffusion Posterior Sampling}
\label{alg:online}
\begin{algorithmic}[1]
\REQUIRE Initial ensemble $\{\mathbf{x}_{0,i}\}_{i=1}^N$, encoder $\psi_\phi$, score net $s_\theta$, drift $f$, diffusion $g$, observations $\{\mathbf{y}_k\}_{k=1}^K$, step $\Delta t$, diffusion times $\{t_\ell\}_{\ell=1}^L$, solver step $\delta t$
\ENSURE Posterior ensembles $\{\{\mathbf{x}_{k,i}\}_{i=1}^N\}_{k=1}^K$ and estimates $\{\hat{\mathbf{x}}_k\}_{k=1}^K$

\STATE $\{\mathbf{x}_{0,i}\}_{i=1}^N \gets$ initial ensemble
\FOR{$k=0$ to $K-1$}
  \STATE \textbf{Propagate step:}
  \FOR{$i=1$ to $N$}
    \STATE Sample $\epsilon_{k,i}\sim\mathcal{N}(0,I)$
    \STATE $\displaystyle
      \mathbf{x}^-_{k+1,i}
      \gets \mathbf{x}_{k,i}
         + f(\mathbf{x}_{k,i},t_k)\,\Delta t
         + g(\mathbf{x}_{k,i},t_k)\,\sqrt{\Delta t}\,\epsilon_{k,i}$
      \quad(Eq.~\eqref{eqn:euler_maruyama})
  \ENDFOR
  \STATE $\mathbf{c}_{k+1}\gets \psi_\phi(\{\mathbf{x}^-_{k+1,i}\}_{i=1}^N)$
      \quad(Eq.~\eqref{eqn:set_encoder})
  \STATE \textbf{Conditioning step:}
  \STATE Initialize $x_i^{(0)}\sim\mathcal{N}(0,\sigma_{t_1}^2I)$,\quad $\sigma_{t_1},\alpha_{t_1}$ via Eq.~\eqref{eqn:ab}
  \FOR{$\ell=1$ to $L$}
    \STATE Compute $\alpha_{t_\ell},\sigma_{t_\ell}$ via Eq.~\eqref{eqn:ab}
    \STATE Sample $\epsilon_i\sim\mathcal{N}(0,I)$
    \STATE $\displaystyle
      s_i = s_\theta(x_i^{(\ell-1)},t_\ell,\mathbf{c}_{k+1}),\quad
      \hat x_{0,i}=\frac{x_i^{(\ell-1)}+\sigma_{t_\ell}^2\,s_i}{\alpha_{t_\ell}}$
      \quad(Eq.~\eqref{eqn:tweedies})
    \STATE $\displaystyle
      \tilde s_i
      = s_i + \mathbf{1}_{\{k>0\}}\,
        \nabla_{\tilde x}\log p(\mathbf{y}_{k+1}\mid \hat x_{0,i})$
    \STATE $\displaystyle
      x_i^{(\ell)}
      = x_i^{(\ell-1)}
        + \bigl[f(x_i^{(\ell-1)},t_\ell)
        - g(t_\ell)^2\,\tilde s_i\bigr]\,
        \delta t
        + g(t_\ell)\,\sqrt{\delta t}\,\epsilon_i$
      \quad(Eq.~\eqref{eqn:reverse_conditional_sde})
  \ENDFOR
  \STATE $\mathbf{x}_{k+1,i}\gets x_i^{(L)}\quad\forall i$
  \STATE $\hat{\mathbf{x}}_{k+1}\gets \frac{1}{N}\sum_{i=1}^N\mathbf{x}_{k+1,i}$
\ENDFOR
\end{algorithmic}
\end{algorithm}

\section{Numerical Experiments}
In this section, we assess the performance of the proposed CSF method on a suite of nonlinear filtering benchmarks that span both low- and high-dimensional regimes.
\subsection{Experiment Setup}
In the offline stage, we employ a standard sequential importance resampling particle filter (PF) to generate datasets of prior‐distribution samples for all benchmark problems.  All models are trained using the Adam optimizer with a StepLR learning‐rate scheduler.  In the online stage, our CSF and all baseline methods, PF, ensemble Kalman filter (EnKF), and the score‐based filter(SF), each use \(N=1000\) particles to ensure a fair comparison.  We adopt the variance‐exploding SDE (VESDE) formulation as the denoising dynamics in the diffusion model, with maximum noise scale \(\sigma_{\max}=25.0\), and perform 500 reverse diffusion steps in the conditioning procedure.  Implementation details of the neural architectures are provided in the Appendix.  Training and evaluation are conducted on an NVIDIA RTX 4090 GPU and a 64-core AMD Ryzen 9 3950X CPU.  We compare the proposed CSF with the SIR filter, the ensemble Kalman filter, and the score‐based filter (SF)\cite{bao2024score} by computing the root mean squared error (RMSE) over \(K=10\) independent realizations:
\begin{equation}\label{eq:rmse}
\mathrm{RMSE}
= \sqrt{\frac{1}{K\,T} \sum_{k=1}^{K} \sum_{t=1}^{T}
\bigl\lVert \hat{\mathbf{x}}_{t}^{(k)} - \mathbf{x}_{t}^{(k)}\bigr\rVert^2}\,.
\end{equation}
\subsection{Numerical Examples}
\subsubsection{Example 1: Double‐Well/Triple-well Potential Problem}
\begin{figure}[htbp]
    \centering
        \includegraphics[width=\linewidth]{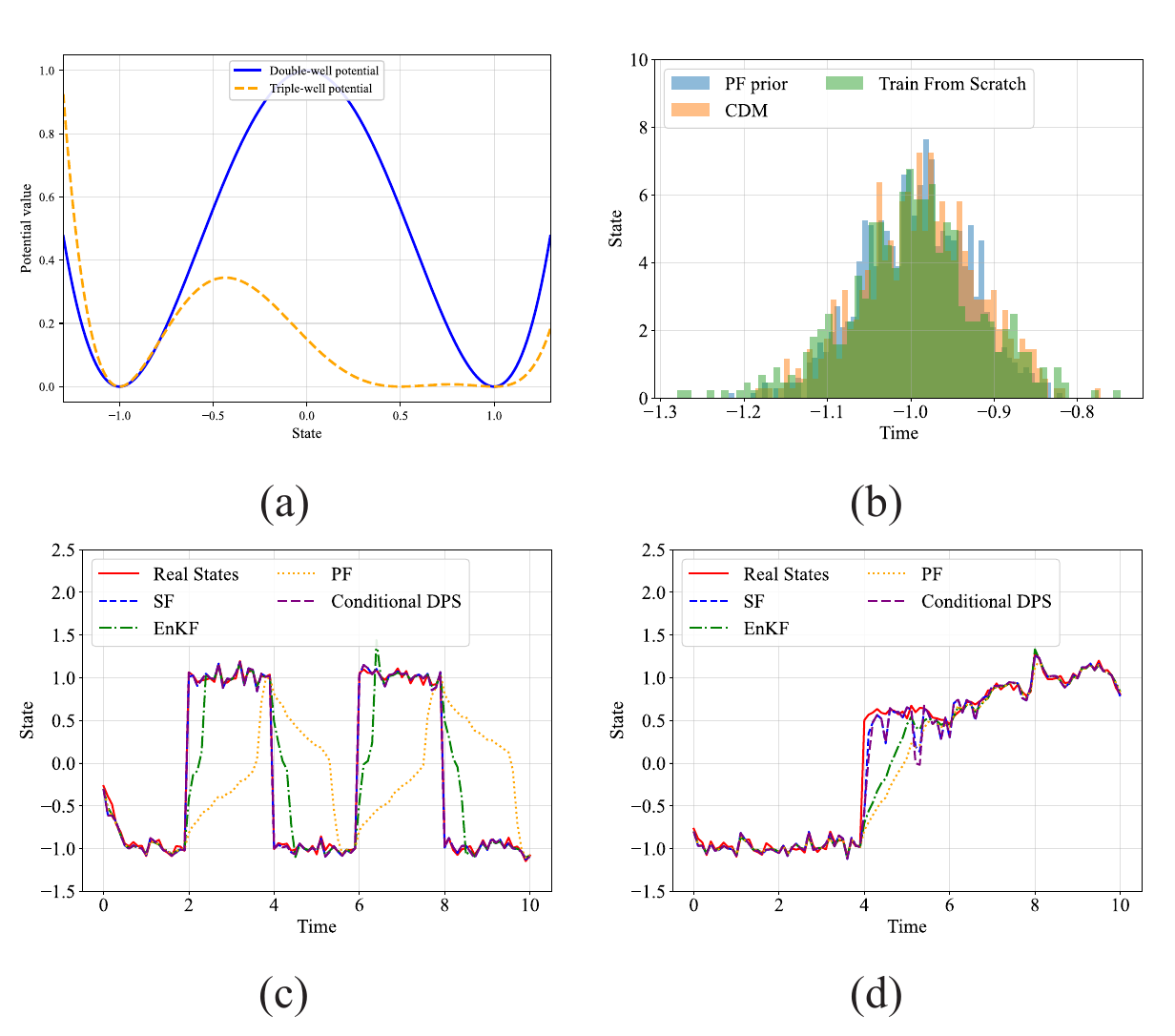}
    \caption{(a) Double/triple-well potential functions.
(b) Comparison of the particle-based input prior with prior samples generated by a diffusion model trained from scratch and by the conditional diffusion model.
(c)  Estimation results of the double-well system using the CSF, SF, PF, and EnKF.
(d) Estimation results of the triple-well system using the CSF, SF, PF, and EnKF.
}
    \label{figure:double_triple}
\end{figure}

We begin with the classic one‐dimensional double‐well/triple-well potential, a benchmark that stresses filters under nonlinear drift and multimodal posterior distributions.  The dynamics and observation model are
\begin{equation}\label{eq:exp1}
\begin{cases}
\mathbf{x}_{k+1} = -\frac{dV}{dx}(\mathbf{x}_k)\Delta t + \sqrt{\Delta t}\,\mathbf{v}_k,\\[6pt]
\mathbf{y}_{k+1} = \mathbf{x}_{k+1}^3 + \mathbf{w}_k,
\end{cases}
\end{equation}
where $\Delta t = 0.1$, \(\mathbf{v}_k\sim \mathcal{N}(0,0.2^2)\) and \(\mathbf{w}_k\sim \mathcal{N}(0,0.1^2)\) are independent standard Gaussian noise. The total length of the trajectory is $K = 100$.  As shown in Fig~\ref{figure:double_triple}(a), the drift term in the double-well and triple-well potential problems is defined as
\begin{equation}
\label{eq:double well}
\left\{
\begin{array}{ll}
V_{\text{double}}(x) &= C(x-1)^{2}(x+1)^{2}, \\
V_{\text{triple}}(x) &= C(x-1)^{2}(x+1)^{2}(x-0.5)^{2},
\end{array}
\right.
\end{equation}
where $C$ is a scaling constant introduced to adjust the magnitude of the drift term. The double-well potential induces two stable minima located near $x=\pm 1$, while the triple-well potential gives rise to three stable wells centered at $x\in\{-1,\,0.5,\,1\}$.  In this setting, it is relatively straightforward to track the state as long as it remains within a single potential well. The main challenge arises when the state unexpectedly transitions from one well to another, leading to a pronounced discrepancy between the observational data and the prior distribution.

In this experiment, we ran a standard PF with \(N=1000\) particles over \(M=1000\) independent trajectories of the double-well potential problem, and at each time step collected the propagated prior ensemble \(\{\mathbf{x}_{t,i}\}_{i=1}^N\) as our offline training dataset.

Figure~\ref{figure:double_triple}(b) illustrates, for a representative prior distribution in the dataset, the inferred sample distributions produced by the pretrained conditional diffusion model and by a score-based diffusion model trained from scratch on this single prior. The results in demonstrate that our method accurately learns the prior information in the filtering problem, providing score approximations of comparable quality to those obtained by directly training a diffusion model. For the double-well potential problem, both our method and the score-based filter (SF) are capable of rapidly capturing state transitions, as score-based posterior sampling exploits the likelihood information more effectively. By contrast, the limited distributional support of the PF and EnKF results in inaccurate tail approximations of the prior distribution, thereby yielding poor posterior estimates. Furthermore, when applying the CSF model trained on the double-well dataset directly to the triple-well potential problem, we still obtain results superior to traditional methods and comparable in accuracy to SF, while achieving an five-fold increase in computational efficiency relative to SF.

\subsubsection{Example 2: Nonlinear filtering with non-Gaussian noise}
We next evaluate the performance of CSF in the presence of heavy-tailed measurement noise by considering a five-dimensional cubic-sensor model with Cauchy-type observations. Specifically, we study the non-linear filtering system
\begin{equation}\label{eq:exp2}
\begin{cases}
\mathbf{x}_{k+1} \;=\; \mathbf{x}_{k} + \cos(\mathbf{x}_{k})\,\Delta t \;+\; \mathbf{v}_{k}, \\[6pt]
\mathbf{y}_{k+1} \;=\; \mathbf{x}_{k+1}^{\,3} \;+\; \mathbf{w}_{k},
\end{cases}
\end{equation}
where the state dimension is $d=5$, the discretization step is $\Delta t=0.01$, and the trajectory length is $K=200$. The process noise is distributed as $\mathbf{v}_{k}\sim\mathcal{N}(\mathbf{0},\,0.1^2\mathbf{I}_d)$, while the observation noise has i.i.d.\ entries $\mathbf{w}_{k}\sim\text{Cauchy}(0,\gamma)$ with $\gamma=0.0544$, obtained via a quantile-based approximation of a unit-variance Gaussian distribution.  To further increase the difficulty of the tracking task, we introduce a random shock to simulate abrupt external perturbations that may arise in practice.

The dataset is generated by running a standard particle filter with $N=1000$ particles over $M=1000$ independent trajectories. During the online stage, the EnKF employs the quantile-based Gaussian approximation to model the likelihood. In this setting, the pretrained conditional diffusion model must capture the heavy-tailed characteristics of different prior distributions and generalize effectively to provide accurate score approximations for the priors encountered in the online stage.
\begin{figure}[htbp]
    \centering
        \includegraphics[width=\linewidth]{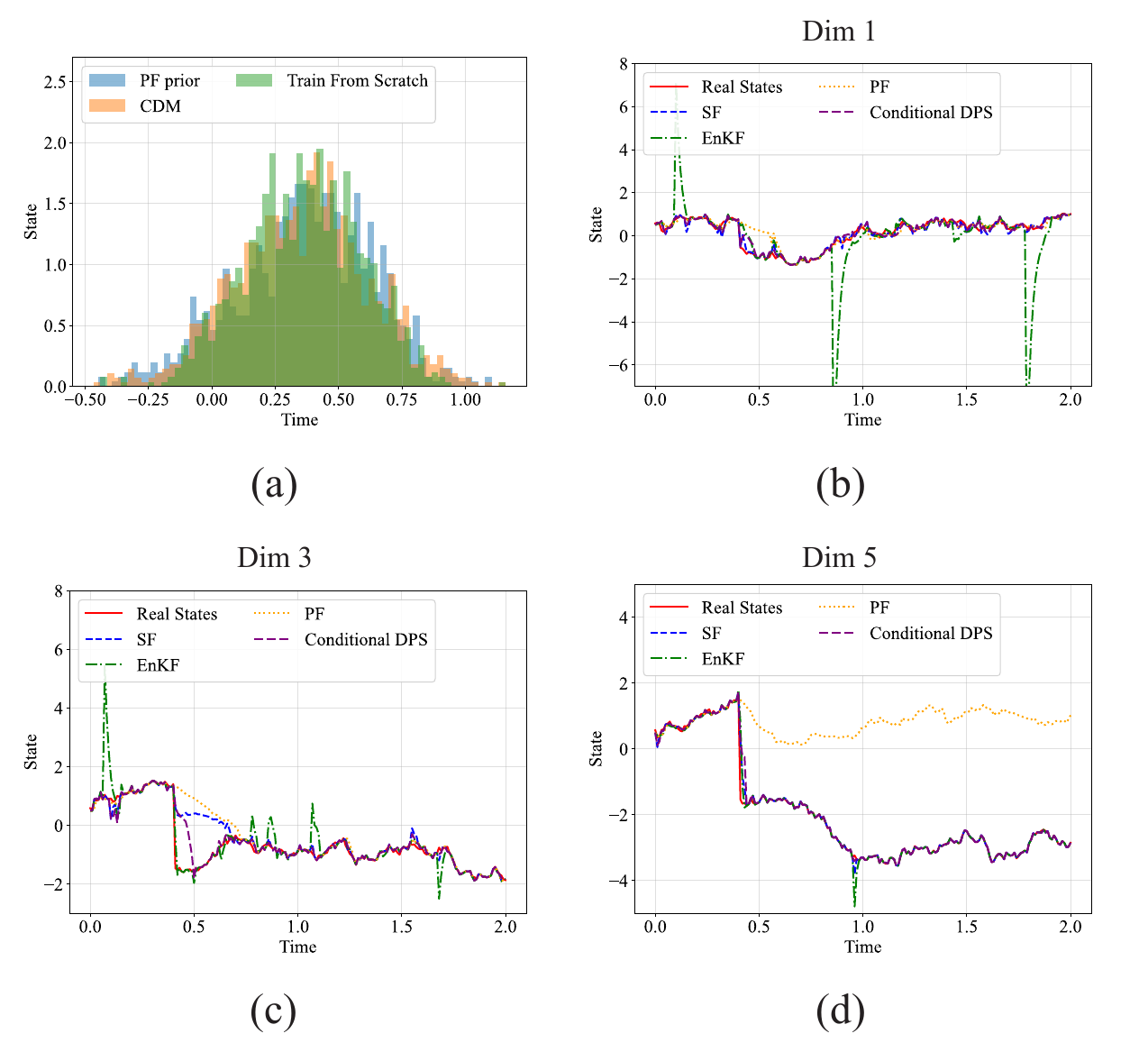}
    \caption{(a) Comparison of the particle-based input prior with prior samples generated by a diffusion model trained from scratch and by the conditional diffusion model(Dimension 1).
(b-d) Estimation results of the cubic sensor filtering system using the CSF, SF, PF, and EnKF.
}
    \label{figure:cauchy}
\end{figure}

Figure~\ref{figure:cauchy}(a) shows that the conditional model, pretrained on a broad collection of heavy-tailed priors, provides a markedly better approximation of the prior distribution than a diffusion model trained on a single distribution. Figures~\ref{figure:cauchy}(b–d) present the estimation results for the components $x_1$, $x_3$, and $x_5$. The proposed CSF consistently outperforms SF, a result further corroborated by Figure~\ref{figure:rmse}. This advantage arises because, in this setting, the prior distributions exhibit heavy-tailed characteristics. Training a diffusion model from scratch within limited computational budgets often fails to capture such complex features, whereas the pretrained CSF, trained extensively in the offline stage, is able to represent these distributions with high fidelity. By contrast, the traditional PF, although initially effective, is unable to track the state dynamics after random shocks. Similarly, the EnKF, which relies on Gaussian approximations, cannot adequately model the posterior under Cauchy noise and eventually breaks down due to the insufficiency of effective samples.

\subsubsection{Example 3: High‐Dimensional Cubic Sensor Problem}
Finally, we tackle a high‐dimensional cubic sensor problem to demonstrate scalability and robustness in large‐scale settings with both independent and correlated state dimensions.  We consider the cubic sensor filtering system in \(\mathbb{R}^n\):
\begin{equation}\label{eq:exp3}
\begin{cases}
\mathbf{x}_{k+1} \;=\; \mathbf{x}_{k} + f(\mathbf{x}_{k})\,\Delta t \;+\; \mathbf{v}_{k}, \\[6pt]
\mathbf{y}_{k+1} \;=\; \mathbf{x}_{k+1}^{\,3} \;+\; \mathbf{w}_{k},
\end{cases}
\end{equation}
with
\[
f(\mathbf{x}) =
\begin{cases}
\cos(\mathbf{x}), & \text{(independent)},\\[4pt]
A_n\,\mathbf{x} + \cos(\mathbf{x}), & \text{(correlated)},
\end{cases}
\quad
A_n = [a_{ij}],\;
a_{ij} = \begin{cases}
0.1, & i+1 = j,\\
-0.5,& i=j,\\
0,   & \text{otherwise},
\end{cases}
\]
where $\Delta t = 0.01$, the process noise is given by $\mathbf{v}_k \sim \mathcal{N}(0,0.1^2)$ and the observation noise by $\mathbf{w}_k \sim \mathcal{N}(0,0.1^2)$, both independent Gaussian perturbations. The trajectory length is set to $K=200$. As in the previous experiments, two random shocks are introduced to emulate external disturbances that may arise in practice. The dataset is generated by running a standard particle filter with $N=1000$ particles over $M=2000$ independent trajectories.

\begin{figure}[htbp]
    \centering
        \includegraphics[width=\linewidth]{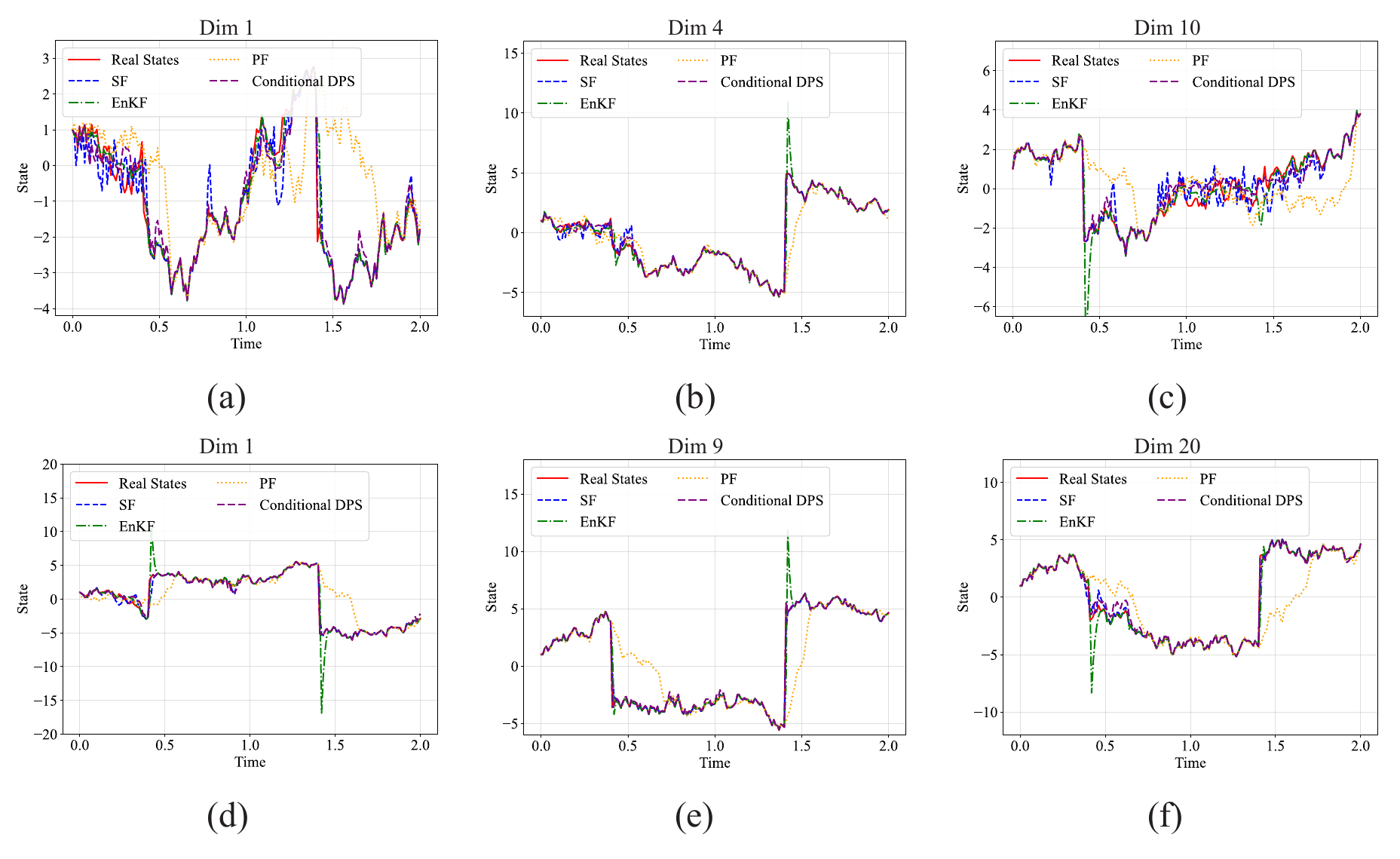}
    \caption{(a)-(c) Estimation results
of the 10D cubic sensor problem with dimension-correlated drift term using the CSF, SF, PF, and EnKF. (d)-(f) Estimation results
of the 20D cubic sensor problem with dimension-correlated drift term using the CSF, SF, PF, and EnKF.
}
    \label{figure:corr}
\end{figure}

\begin{figure}[htbp]
    \centering
        \includegraphics[width=\linewidth]{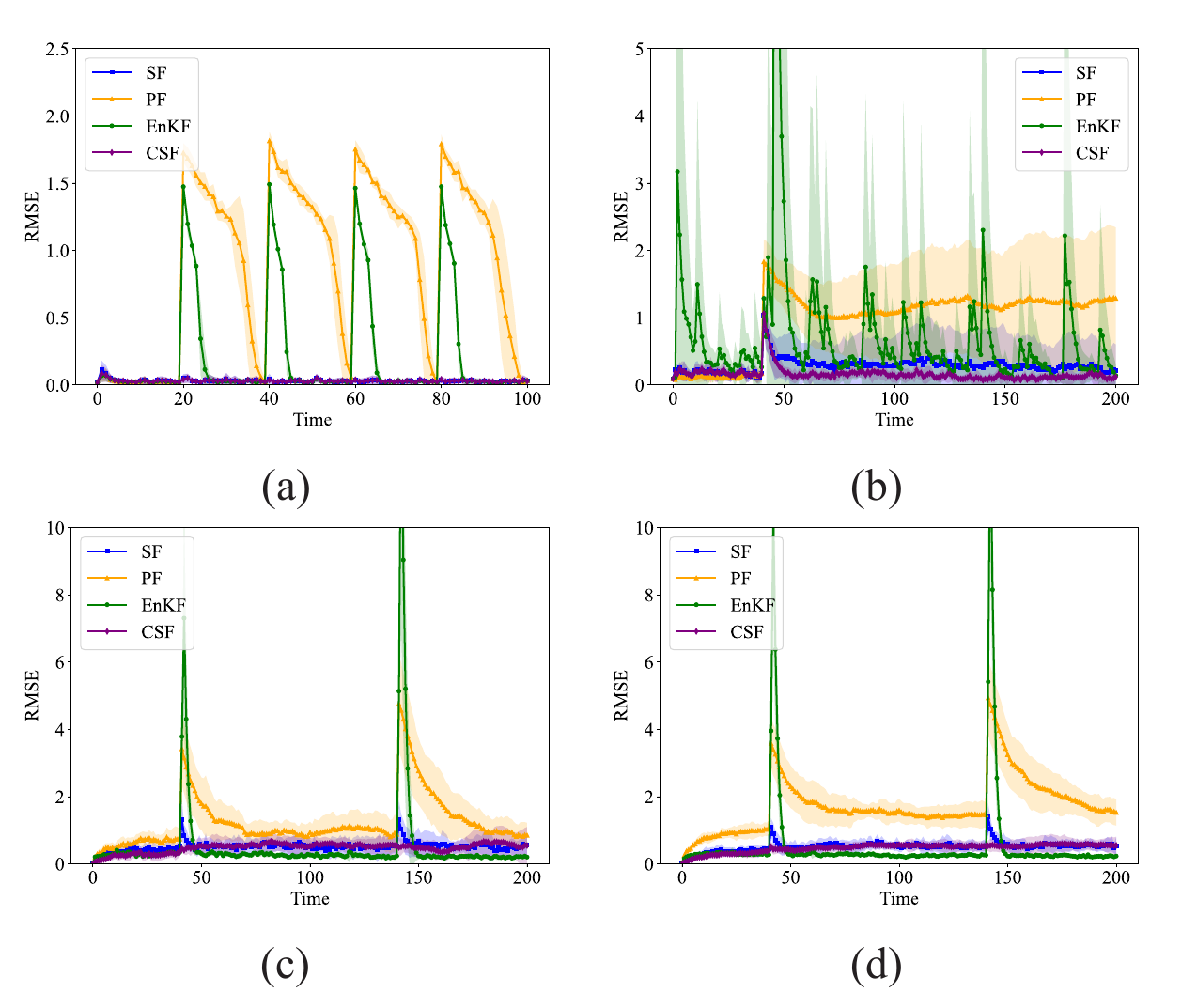}
\caption{Average RMSEs for (a) the double-well potential problem; (b) the cubic-sensor problem with Cauchy-type noise; and (c)–(d) the 10D and 20D cubic-sensor problems with dimension-correlated drift terms.}
    \label{figure:rmse}
\end{figure}

The numerical results for the 10D and 20D problems with dimension-correlated drift terms are reported in Fig.~\ref{figure:corr} and Table~\ref{tab:rmse-by-task}. Our method exhibits superior performance in both cases, accurately tracking the evolution of the high-dimensional states even in the presence of unexpected shocks. In contrast, the SF estimates are less stable, as the continual evolution of high-dimensional priors requires substantial computational effort to model accurately at certain time steps—a difficulty that our approach mitigates through large-scale offline pretraining. For these problems, PF performs significantly worse than the score-based methods due to the curse of dimensionality, which necessitates prohibitively large sample sizes. As shown in Fig.~\ref{figure:rmse}, the EnKF achieves slightly better performance than the score-based approaches under nominal conditions; however, once unexpected shocks occur, its estimates diverge, resulting in overall errors higher than those of both CSF and SF. This highlights the robustness of CSF, as further confirmed by the results in Table~\ref{tab:rmse-by-task}.

\begin{figure}[htbp]
    \centering
        \includegraphics[width=\linewidth]{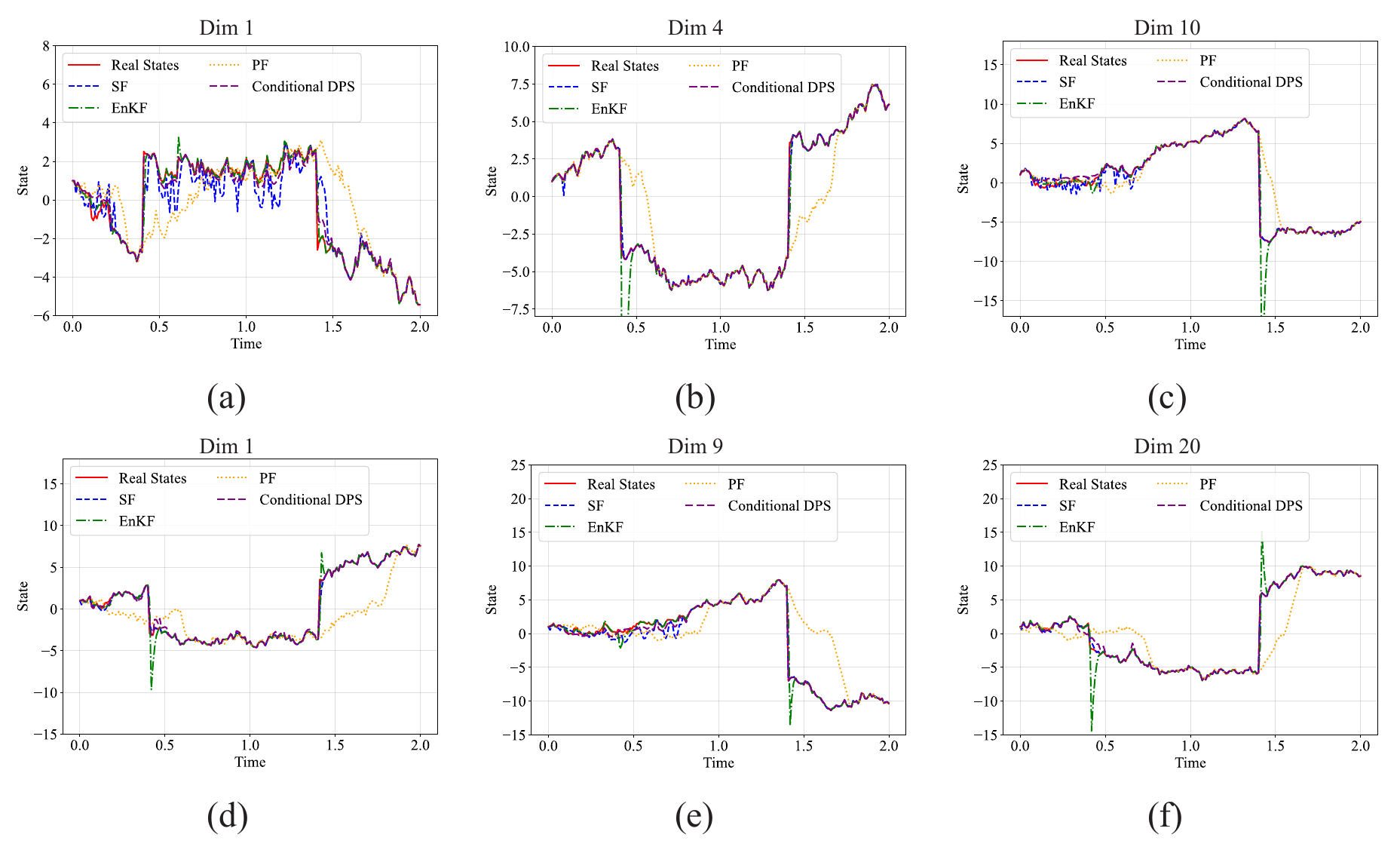}
    \caption{(a)-(c) Estimation results
of the 10D cubic sensor problem with dimension-independent drift term using the CSF, SF, PF, and EnKF. (d)-(f) Estimation results
of the 20D cubic sensor problem with dimension-independent drift term using the CSF, SF, PF, and EnKF.
}
    \label{figure:indep}
\end{figure}

Figure~\ref{figure:indep} and Table~\ref{tab:rmse-by-task} present the estimation results for the 10D and 20D problems with dimension-independent drift terms. This setting is more challenging, as the evolution across dimensions is mutually independent, yielding more complex prior and posterior distributions that require larger particle sets for accurate approximation. In this case, our method achieves overall superior performance compared with SF and other baselines, exhibiting behavior consistent with the dimension-correlated examples. Importantly, during the online stage, our approach requires no additional training and attains a fivefold speedup using the standard DDPM sampler (see Table~\ref{tab:time-by-task}). This suggests that even greater acceleration can be expected when more advanced sampling techniques are employed.

\begin{table}[htbp]
  \centering
  \label{tab:rmse-by-task}
  \small
  \setlength{\tabcolsep}{4pt} % 缩小列间距
  \begin{tabular}{lccccccc}
    \toprule
    \textbf{Model} & \textbf{Double Well} & \textbf{Triple Well} & \textbf{Cauchy} & \textbf{10D \textit{I}} & \textbf{10D \textit{C}} & \textbf{20D \textit{I}} & \textbf{20D \textit{C}} \\
    \midrule
    CSF  & \textbf{0.038} & 0.185 & \textbf{0.217} & \textbf{0.748} & \textbf{0.627}  & \textbf{0.764} &\textbf{0.5978} \\
    SF         & 0.040 & \textbf{0.176} & 0.285 & 0.750 & 0.667  & 0.868 & 0.661\\
    EnKF       & 0.474 & 0.249 & 2.418 & 1.481 & 1.108 & 1.338 & 1.178\\
    PF         & 1.117 & 0.317 & 1.162 & 1.928 & 1.489 & 2.534 &1.937\\
    \bottomrule
  \end{tabular}
    \caption{%
    Average RMSEs across different tasks using CSF, SF, EnKF, and PF, computed over 10 independent runs. 
    10D/20D I denotes the 10- and 20-dimensional problems with dimension-independent drift terms, while 
    10D/20D C refers to the corresponding problems with dimension-correlated drift terms.}
\end{table}

\section{Conclusions and Discussions}
In this paper, we introduced the conditional score-based filter (CSF) for general nonlinear filtering problems. The proposed framework leverages a set-transformer prior encoder to extract essential statistical information from prior ensembles across different dimensions and noise types. This enables the conditional diffusion model to accurately represent the evolving prior distributions inherent in nonlinear filtering. Combined with score-based posterior sampling, the CSF method demonstrates strong performance in challenging scenarios, including filtering under heavy-tailed measurement noise, high-dimensional nonlinear dynamics, and unexpected external shocks. 

Methodologically, our approach disentangles the modeling of prior distributions from the posterior sampling stage in score-based filtering. This separation not only improves computational efficiency by an order of magnitude but also highlights the ability of the conditional diffusion model to generalize across problems, transferring knowledge gained from one setting to another. 

Overall, our study illustrates the potential of score-based filtering to deliver accurate and stable state estimates in complex, high-dimensional nonlinear systems. While this work focused on diffusion posterior sampling (DPS), which provides an approximate score-based posterior sampler, future research may incorporate more accurate posterior sampling schemes as well as accelerated algorithms to further enhance performance. Another important direction is to establish a comprehensive convergence theory for score-based filtering under different posterior sampling strategies.

\begin{table}[htbp]
  \centering
  \label{tab:time-by-task}
  \small
  \setlength{\tabcolsep}{4pt} % 缩小列间距
  \begin{tabular}{lcccc}
    \toprule
    \textbf{Time/s} & \textbf{Double Well}  & \textbf{Cauchy}  & \textbf{10D}  & \textbf{20D }  \\
    \midrule
    CSF  & \textbf{1.25} & \textbf{1.55} & \textbf{1.61} & \textbf{1.65}  \\
    SF         & 5.85  & 7.88 & 7.95 & 8.06\\
    \bottomrule
  \end{tabular}
    \caption{%
   Comparison of computational time for each step between different CSF and SF across tasks.}
\end{table}

% \section*{Acknowledgments}
% We would like to acknowledge the assistance of volunteers in putting
% together this example manuscript and supplement.

\bibliographystyle{siamplain}
\bibliography{references}

\end{document}

%% file: ex_shared.tex
\usepackage{lipsum}
\usepackage{amsfonts}
\usepackage{graphicx}
\usepackage{epstopdf}
\usepackage{booktabs}
\usepackage{algorithmic}
\ifpdf
  \DeclareGraphicsExtensions{.eps,.pdf,.png,.jpg}
\else
  \DeclareGraphicsExtensions{.eps}
\fi

\newsiamremark{remark}{Remark}
\newsiamremark{hypothesis}{Hypothesis}
\crefname{hypothesis}{Hypothesis}{Hypotheses}
\newsiamthm{claim}{Claim}
\newsiamremark{fact}{Fact}
\crefname{fact}{Fact}{Facts}

\headers{Efficient Conditional Score-based Filter}{Zhijun Zeng, Weiye Gan, Junqing Chen, Zuoqiang Shi.}

\title{ An Efficient Conditional Score-based Filter for High Dimensional Nonlinear Filtering Problems\thanks{Submitted to the editors DATE.
\funding{This work was supported by National Natural Science Foundation of China under grant 12071244.}}}

\author{Zhijun Zeng\thanks{Department of Mathematical Sciences, Tsinghua University, Beijing, China 
  (\email{zengzj22@mails.tsinghua.edu.cn}).}\and
Weiye Gan\thanks{Department of Mathematical Sciences, Tsinghua University, Beijing, China 
  (\email{gwy22@mails.tsinghua.edu.cn}).}
 \and Junqing Chen\thanks{Department of Mathematical Sciences, Tsinghua University, Beijing, China  
  (\email{jqchen@tsinghua.edu.cn}).}
\and Zuoqiang Shi\thanks{Yau Mathematical Sciences Center, Tsinghua University, Beijing, China and Yanqi Lake Beijing Institute of Mathematical Sciences and Applications, Beijing, China  
  (\email{zqshi@tsinghua.edu.cn}).Corresponding author.}}
\usepackage{amsopn}